\documentclass[9pt,conference]{IEEEtran}
\ifCLASSINFOpdf
\usepackage{graphicx}
\graphicspath{{images/},{prebuiltimages/}}
\else
\fi
\usepackage{xy}
\usepackage{times}
\usepackage{subfigure}
\usepackage{mystyle}
\usepackage{amsmath,amsthm}
\usepackage{amsopn,amssymb}
\usepackage{bbm,dsfont}
\usepackage{mathrsfs}
\usepackage[utf8]{inputenc}

\usepackage[textwidth=2.450cm, textsize=tiny]{todonotes}

\begin{document}
%
\title{Characterizing the maximum parameter of the total-variation denoising through the pseudo-inverse of the divergence}

\author{\IEEEauthorblockN{Charles-Alban Deledalle $\quad$ Nicolas Papadakis}
\IEEEauthorblockA{IMB, CNRS, Bordeaux INP\\ Universit\'e Bordeaux, Talence, France\\
Email: firstname.lastname@math.u-bordeaux.fr}
\and
\IEEEauthorblockN{Joseph Salmon}
\IEEEauthorblockA{LTCI, CNRS, T\'el\'ecom ParisTech\\
Université Paris-Saclay, France\\
Email: joseph.salmon@telecom-paristech.fr}
\and
\IEEEauthorblockN{Samuel Vaiter}
\IEEEauthorblockA{IMB, CNRS\\
Universit\'e de Bourgogne, Dijon, France\\
Email: samuel.vaiter@u-bourgogne.fr}
}


%


\maketitle

\begin{abstract}
  We focus on the maximum regularization parameter
  for anisotropic total-variation denoising.
  It corresponds to the minimum value
  of the regularization parameter above which the solution remains constant.
  While this value is well know for the Lasso, such a critical value has not been investigated in details for the total-variation.
  Though, it is of importance when tuning the regularization parameter as
  it allows fixing an upper-bound on the grid for which the optimal parameter is sought.
  We establish a closed form expression
  for the one-dimensional case,
  as well as an upper-bound for the two-dimensional case, that appears reasonably tight in practice.
  This problem is directly linked to
  the computation of the pseudo-inverse of the divergence,
  which can be quickly obtained by performing convolutions in the Fourier domain.
\end{abstract}


%
\IEEEpeerreviewmaketitle

\section{Introduction}

We consider the reconstruction of a $d$-dimensional signal
(in this study $d=1$ or $2$)
from its noisy observation $y \!=\! x + w \!\in\! \RR^n$ with
$w\!\in\! \RR^n$.
Anisotropic TV regularization writes, for $\lambda > 0$, as \cite{rudin1992nonlinear}
\begin{align}
  \label{sub:denoising_case}
  x^\star = \uargmin{x \in \RR^n} \frac{1}{2} \norm{y-x}_2^2 + \lambda \norm{\nabla x}_1
\end{align}
with $\nabla x \!\in\! \RR^{dn}$ being the concatenation of the $d$ components
of the discrete periodical gradient vector field of $x$,
and $\norm{\nabla x}_{1} \!=\! \sum_i |(\nabla x)_i|$ being a sparsity
promoting term. The operator $\nabla$ acts as a convolution
which writes in the one dimensional case ($d=1$)
\begin{align}
\nabla =
F^+ \diag(K_\rightarrow) F
\qandq
\diverg =
F^+ \diag(K_\leftarrow) F
\end{align}
where $\diverg = -\nabla^\top$ (where $^\top$ denotes the adjoint),
$F : \RR^n \mapsto \CC^n$ is the Fourier transform, $F^+ = \text{Re}[F^{-1}]$
is its pseudo-inverse and $K_\rightarrow \in \CC^n$ and $K_\leftarrow \in \CC^n$ are
the Fourier transforms of the kernel functions performing
forward and backward finite differences respectively.
Similarly, we define in the two dimensional case ($d=2$)
\begin{align}
\nabla &=
\begin{pmatrix}
  F^+ & 0\\
  0 & F^+
\end{pmatrix}
\begin{pmatrix} \diag(K_\downarrow) \\ \diag(K_\rightarrow) \end{pmatrix}
F\\
\qandq
\diverg &=
F^+
\begin{pmatrix} \diag(K_\uparrow) & \diag(K_\leftarrow) \end{pmatrix}
\begin{pmatrix}
  F & 0\\
  0 & F
\end{pmatrix}
\end{align}
where $K_\rightarrow \in \CC^n$ and $K_\leftarrow \in \CC^n$
(resp.~$K_\downarrow \in \CC^n$ and $K_\uparrow \in \CC^n$)
perform forward and backward finite difference in the horizontal
(resp.~vertical) direction.

\section{General case}

For the general case, the following proposition provides an expression
of the maximum regularization parameter $\lambda_{\max}$ as the solution
of a convex but non-trivial optimization problem
(direct consequence of the Karush-Khun-Tucker condition).
\begin{prop}
  Define for $y\in \RR^n$,
  \begin{align}\label{eq:lambda_max}
    \lambda_{\max}=\umin{\zeta \in \Ker[\diverg] } \|\diverg^+ y + \zeta \,\|_{\infty}
  \end{align}
  where $\diverg^+$ is the Moore-Penrose pseudo-inverse of $\diverg$
  and $\Ker[\diverg]$ its null space.
  Then,
  $
  x^\star = \frac{1}{n} \One_n \One_n^\top y
  $
  if and only if $\lambda \geq \lambda_{\max}$.
\end{prop}

\section{One dimensional case}

In the 1d case, $\Ker[\diverg] = \Span(\One_n)$ and thus
the optimization problem can be solved by computing $\diverg^+$
in the Fourier domain, in $O(n \log n)$ operations, as shown in the next corollary.

\begin{cor}
For $d=1$,
$
  \lambda_{\max} = \tfrac{1}{2} [\max(\diverg^+y) - \min(\diverg^+y)],
$
\begin{align}
  \qwhereq &\diverg^+= F^+ \diag(K_\uparrow^+) F\\
  \qandq&
  ({K}_\uparrow^+)_i =
  \left\{\begin{array}{ll}
  \frac{(K_\uparrow)^*_i}{|(K_\uparrow)_i|^2}
  & \text{if} \quad |(K_\uparrow)_i|^2 > 0\\
  0 & \text{otherwise}
  \end{array}\right.~,
  \nonumber
\end{align}
and $^*$ denotes the complex conjugate.
\end{cor}
Note that the condition $|(K_\uparrow)_i|^2 > 0$ is satisfied
everywhere except for the zero frequency.
In the non-periodical case, $\diverg$ is the
incidence matrix of a tree whose pseudo-inverse
can be obtained following \cite{bapat1997moore}.

\section{Two dimensional case}

In the 2d case, $\Ker[\diverg]$ is the orthogonal of the vector space of signals satisfying
Kirchhoff's voltage law on all cycles of the periodical grid.
Its dimension is $n+1$. It follows that the optimization problem becomes much harder.
Since our motivation is only to provide an approximation of $\lambda_{\max}$,
we propose to compute an upper-bound in $O(n \log n)$ operations
thanks to the following corollary.
\begin{cor}
For $d=2$,
$
\lambda_{\max}\leq
\underbrace{\tfrac{1}{2} [\max(\diverg^+y) \!-\! \min(\diverg^+y)]}_{\lambda_{\mathrm{bnd}}},
$\vspace{-0.5em}
\begin{align}
  \qwhereq&
  \diverg^+ =
\begin{pmatrix}
  F^+ & 0\\
  0 & F^+
\end{pmatrix}
\begin{pmatrix} \diag(\tilde{K}^+_\uparrow) \\ \diag(\tilde{K}^+_\leftarrow) \end{pmatrix}
F, \qandq\\
(\tilde{K}_\uparrow^+)_i &=
\left\{\begin{array}{ll}
      \frac{(K_\uparrow)^*_i}{|(K_\uparrow)_i|^2 + |(K_\leftarrow)_i|^2}
      & \text{if} \quad |(K_\uparrow)_i|^2 + |(K_\leftarrow)_i|^2 > 0\\
      0 & \text{otherwise}
\end{array}\right.~,
\nonumber\\
(\tilde{K}_\leftarrow^+)_i &=
\left\{\begin{array}{ll}
      \frac{(K_\leftarrow)^*_i}{|(K_\uparrow)_i|^2 + |(K_\leftarrow)_i|^2}
      & \text{if} \quad |(K_\uparrow)_i|^2 + |(K_\leftarrow)_i|^2 > 0\\
      0 & \text{otherwise}
\end{array}\right.~.
\nonumber
\end{align}
\end{cor}
Note that the condition $|(K_\uparrow)_i|^2 + |(K_\leftarrow)_i|^2 > 0$ is again satisfied
everywhere except for the zero frequency.
Remark also that this result can be straightforwardly extended to the case
where $d>2$.

\section{Results and discussion}
\begin{figure*}[!t]
  \centering
  \subfigure[$y$]{\includegraphics[height=0.17\linewidth]{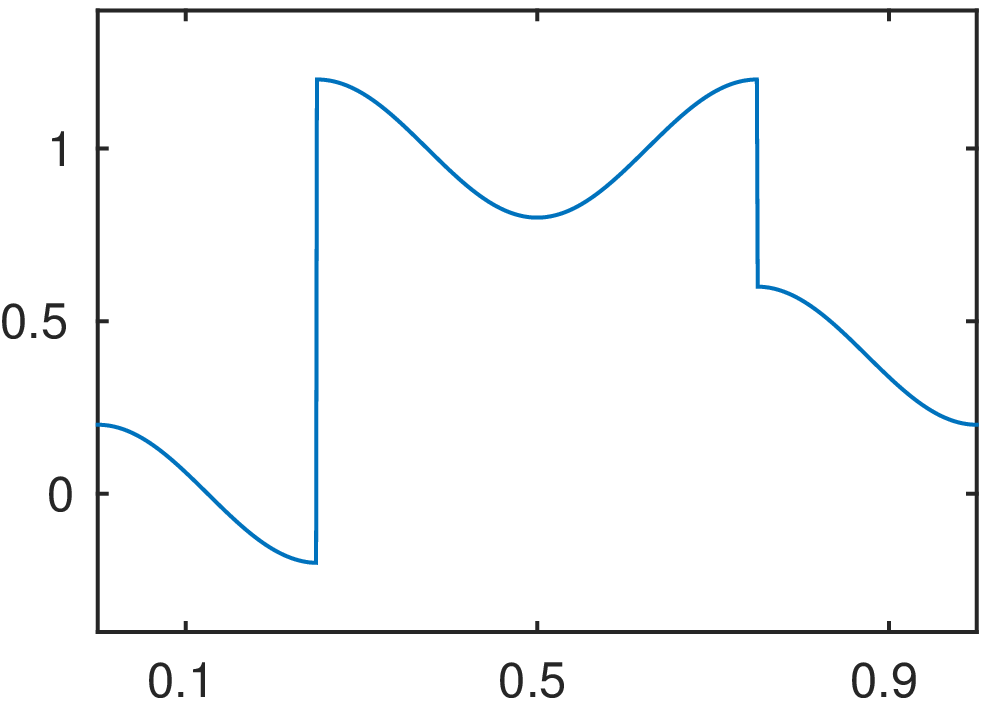}}\hfill%
  \subfigure[$F^+ K^+_\uparrow$]{\includegraphics[height=0.17\linewidth]{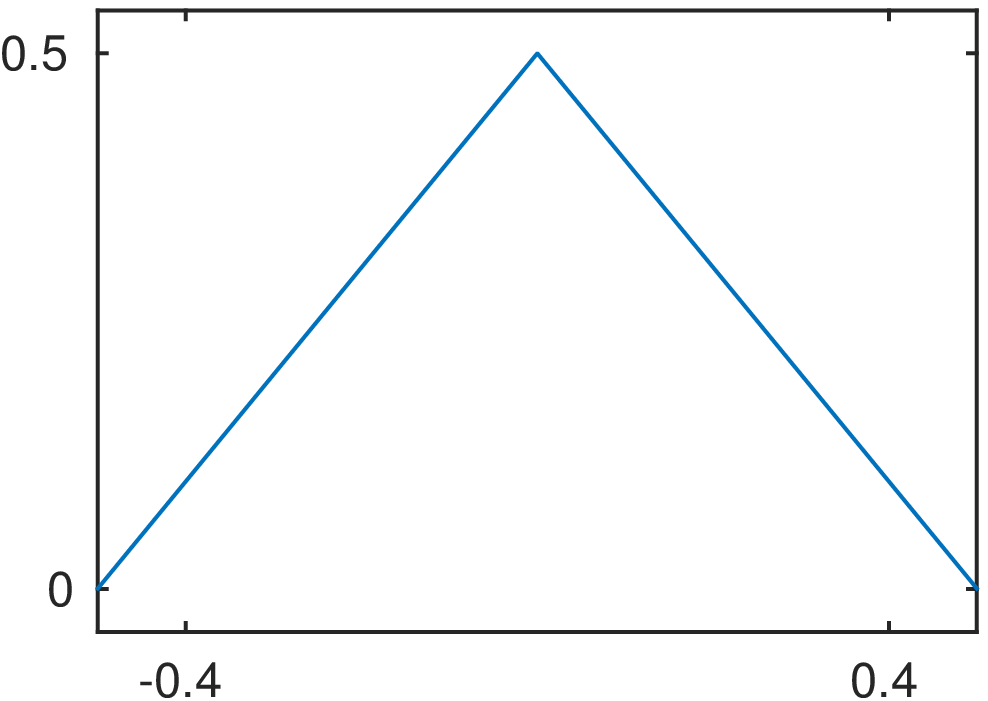}}\hfill%
  \subfigure[$  \diverg^+ y$]{\includegraphics[height=0.17\linewidth]{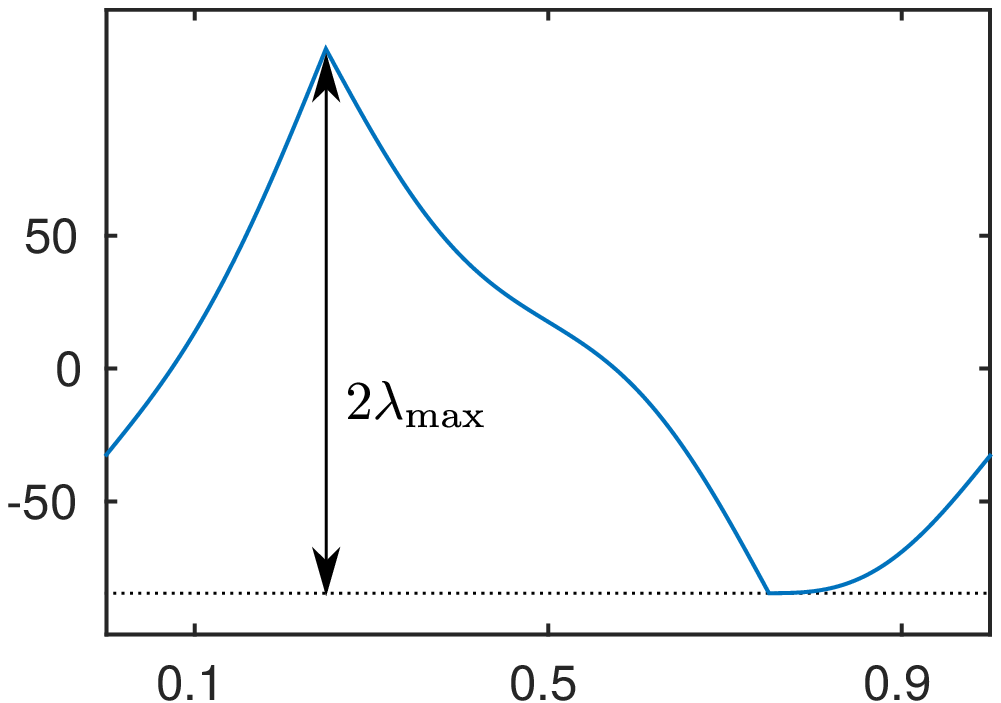}}\hfill%
  \subfigure[$\diverg \diverg^+ y$]{\includegraphics[height=0.17\linewidth]{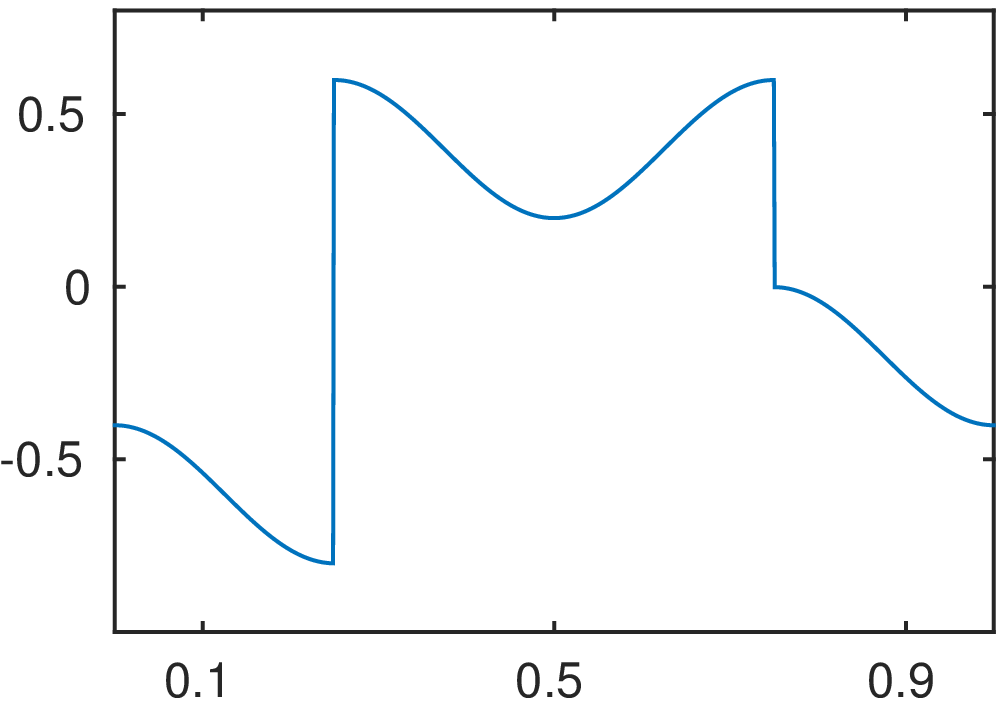}}%
  \caption{(a) A 1d signal $y$. (b) The convolution kernel $F^+ K^+_\uparrow$ that realizes
    the pseudo inversion of the divergence.
    (c) The signal $\diverg^+ y$ on which we can read the value of $\lambda_{\max}$.
    (d) The signal $\diverg \diverg^+ y$ showing that one can reconstruct $y$
    from $\diverg^+ y$ up to its mean component.
  }
  \label{fig:tv_1d_periodic}
  \vspace{1em}
  \centering
  \hspace{0.24\linewidth}\hfill
  \subfigure{\includegraphics[width=0.24\linewidth]{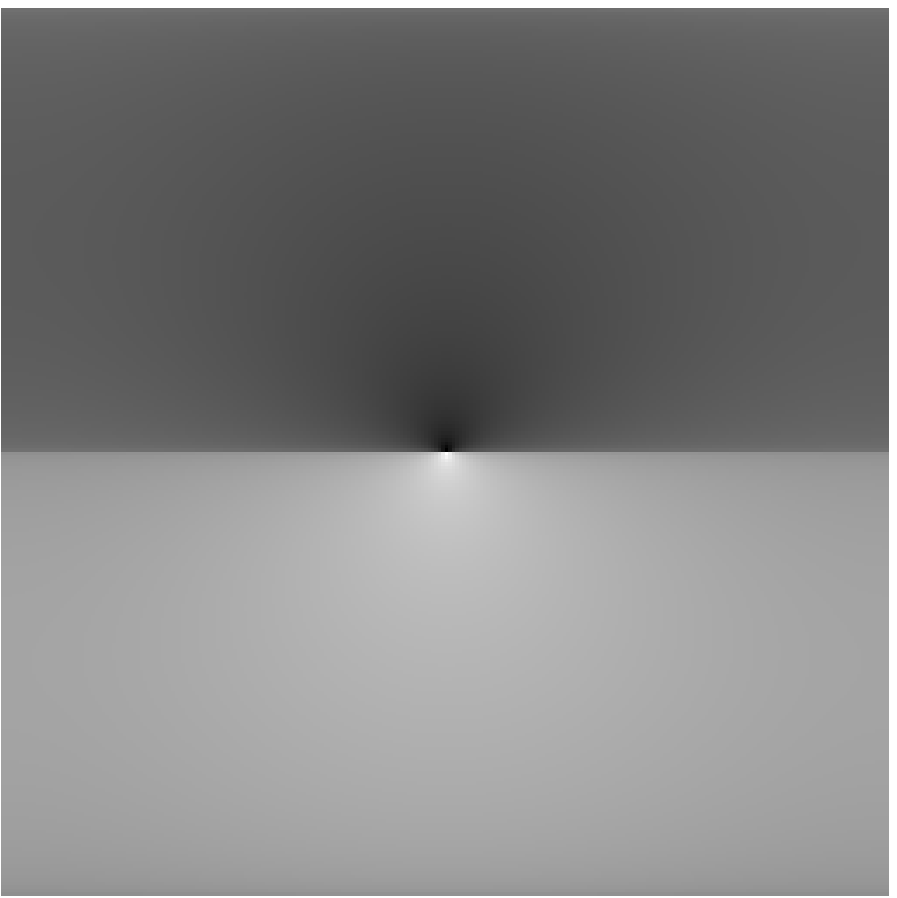}}\hfill%
  \subfigure{\includegraphics[width=0.24\linewidth]{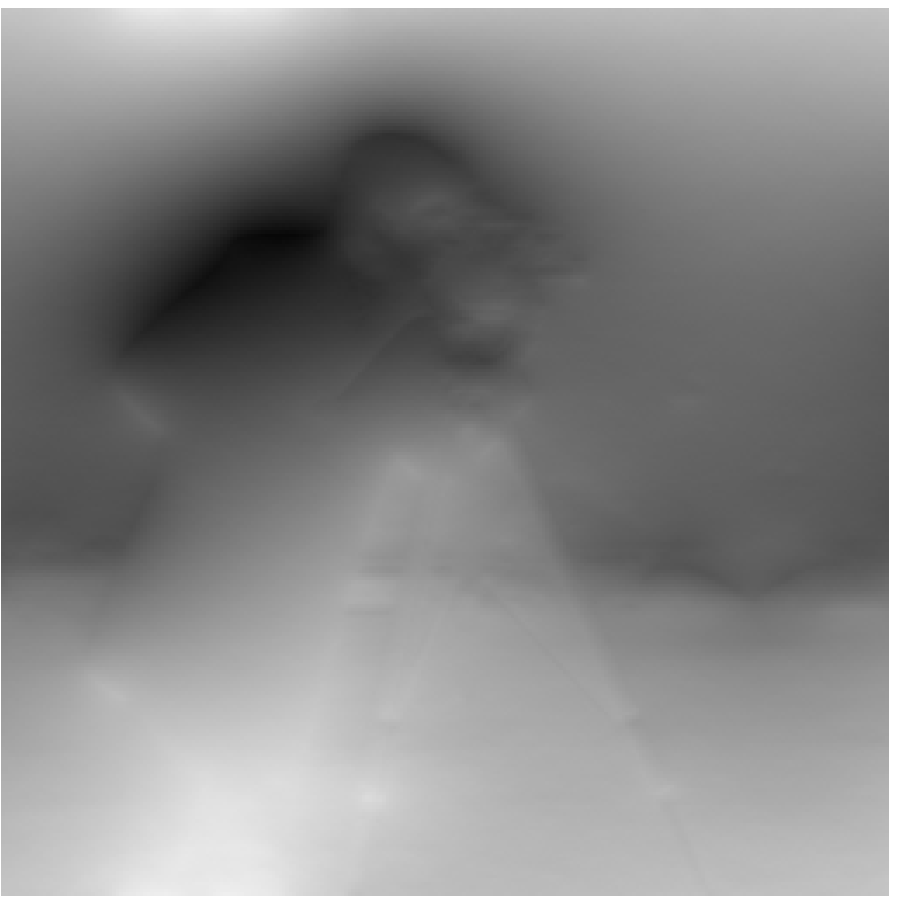}}\hfill%
  \hspace{0.24\linewidth}\hfill{ }\\
  \setcounter{subfigure}{0}%
  \subfigure[$y$ (range {[$0, 255$]})]{\includegraphics[width=0.24\linewidth]{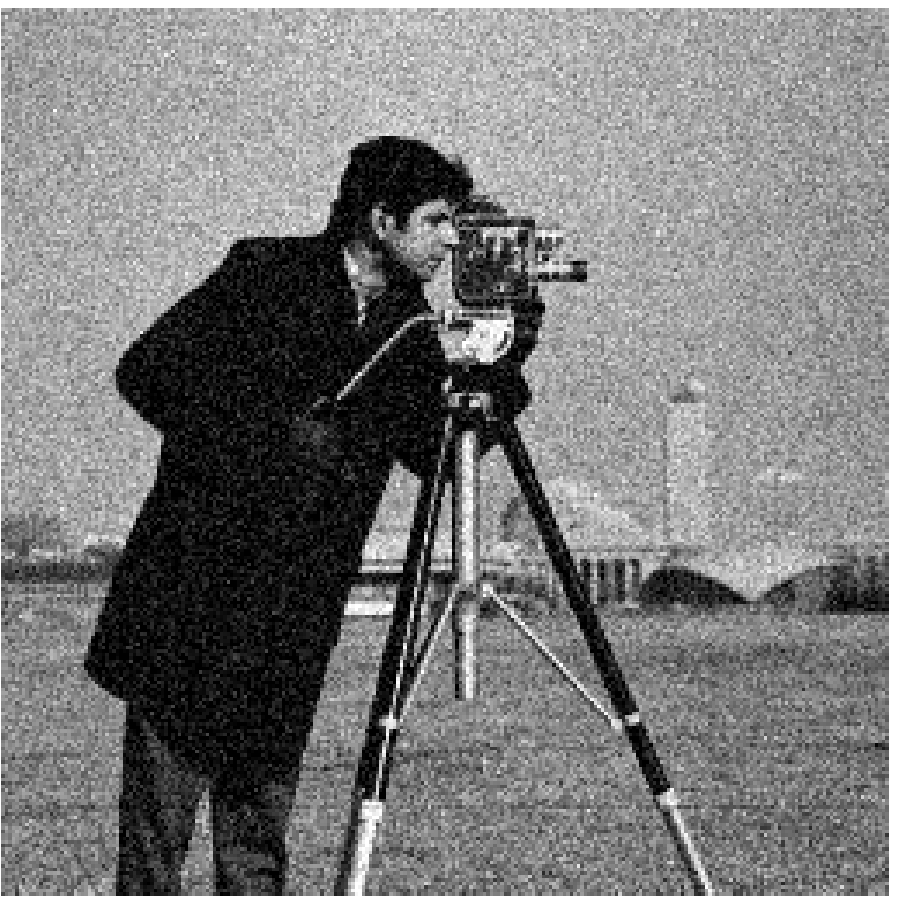}}\hfill%
  \subfigure[$F^+ \tilde{K}^+_\uparrow$, $F^+ \tilde{K}^+_\leftarrow$ (in power scale)]{\includegraphics[width=0.24\linewidth]{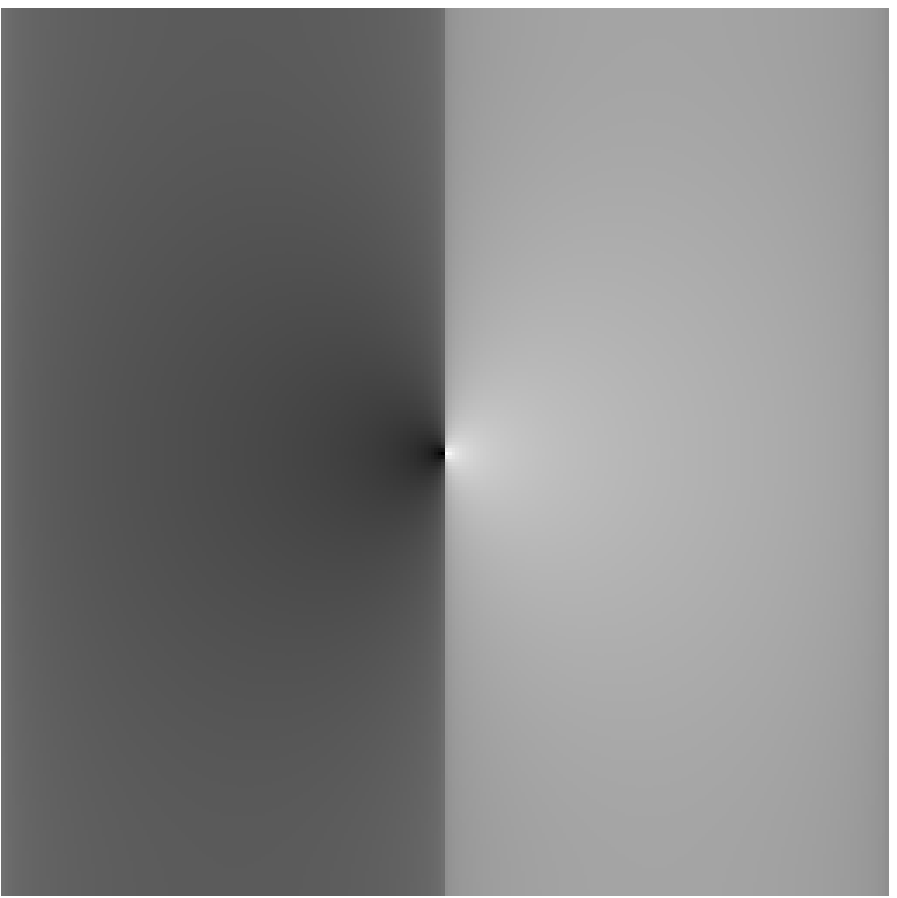}}\hfill%
  \subfigure[$\diverg^+ y$]{\includegraphics[width=0.24\linewidth]{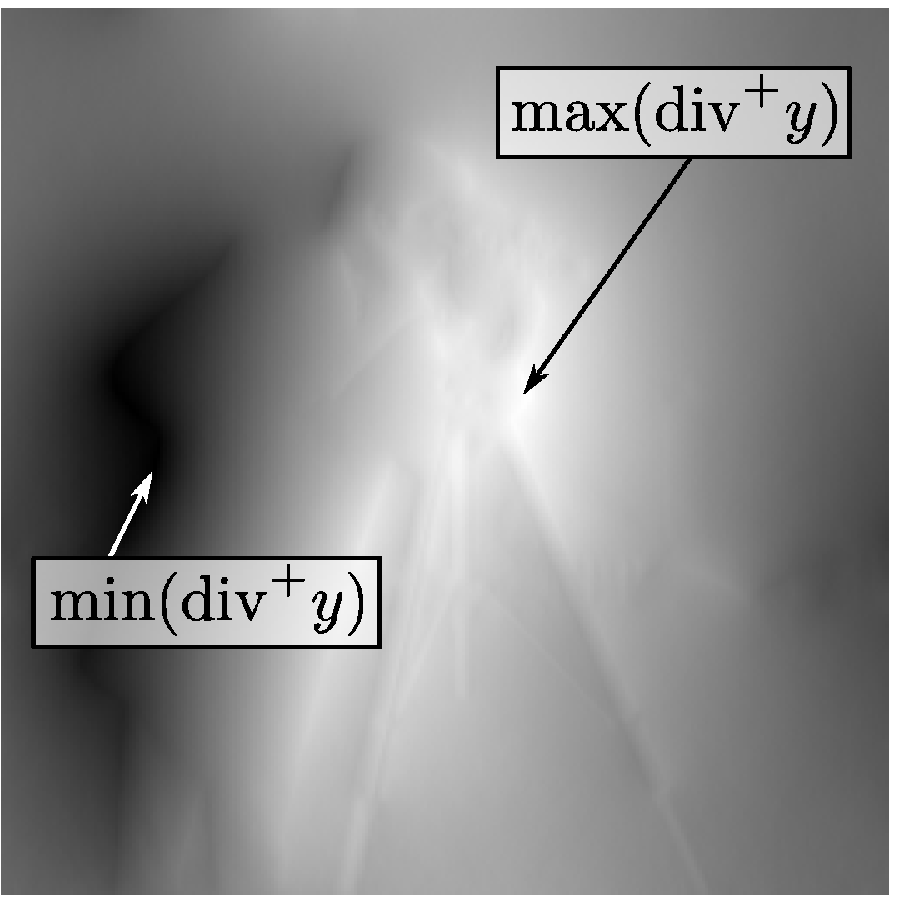}}\hfill%
  \subfigure[$\diverg \diverg^+ y$ (range {[$-119, 136$]})]{\includegraphics[width=0.24\linewidth]{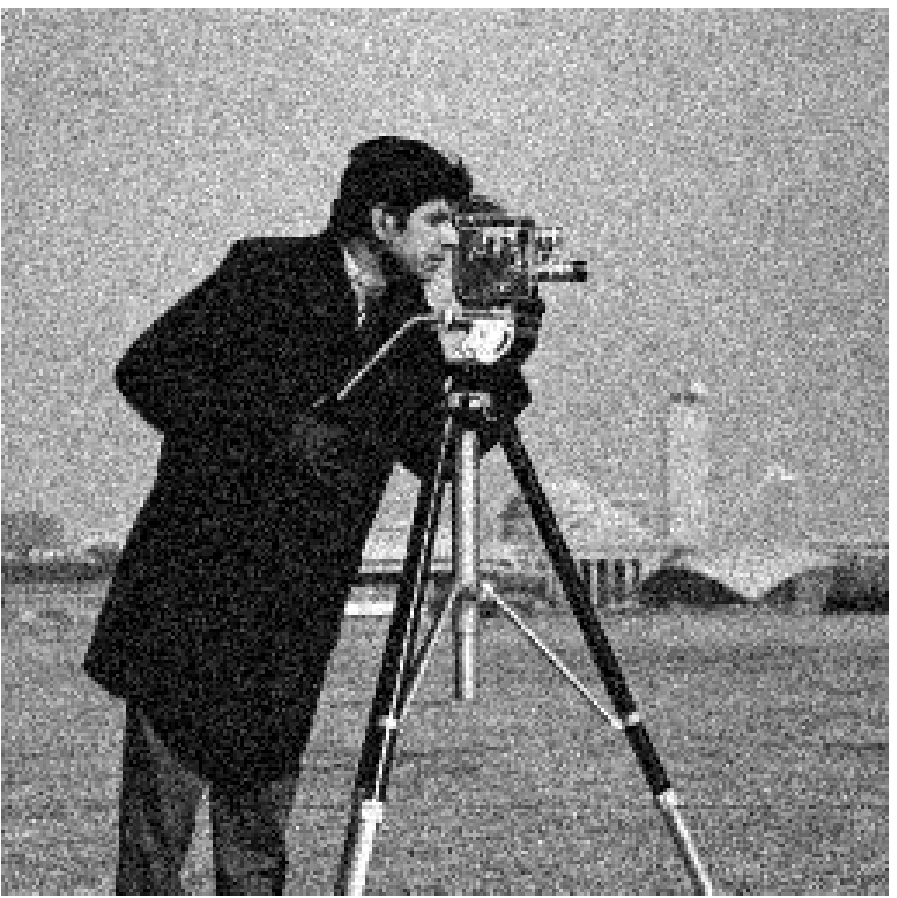}}
  \caption{
    (a) A 2d signal $y$.
    (b) The convolution kernels $F^+ K^+_\uparrow$ and $F^+ \tilde{K}^+_\leftarrow$ that realizes
    the pseudo inversion of the divergence.
    (c) The absolute value of the two coordinates of the vector field $\diverg^+ y$ on which we can read the upper-bound $\lambda_{\mathrm{bnd}}$ of $\lambda_{\max}$.
    (d) The image $\diverg \diverg^+ y$ showing again that one can reconstruct $y$
    from $\diverg^+ y$ up to its mean component.
  }
  \label{fig:tv_2d_periodic}
  \vspace{1em}
  \subfigure[]{\includegraphics[height=0.2\linewidth]{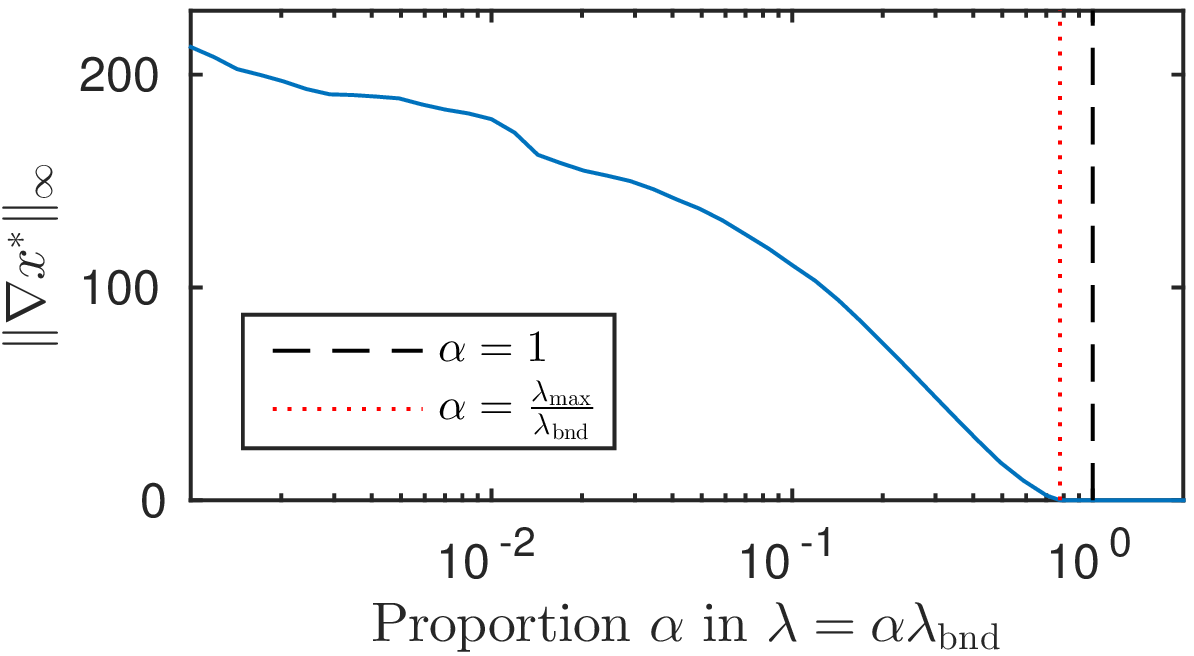}}\hfill
  \subfigure[$\lambda = 10^{-3} \lambda_{\mathrm{bnd}}$]{\includegraphics[height=0.2\linewidth]{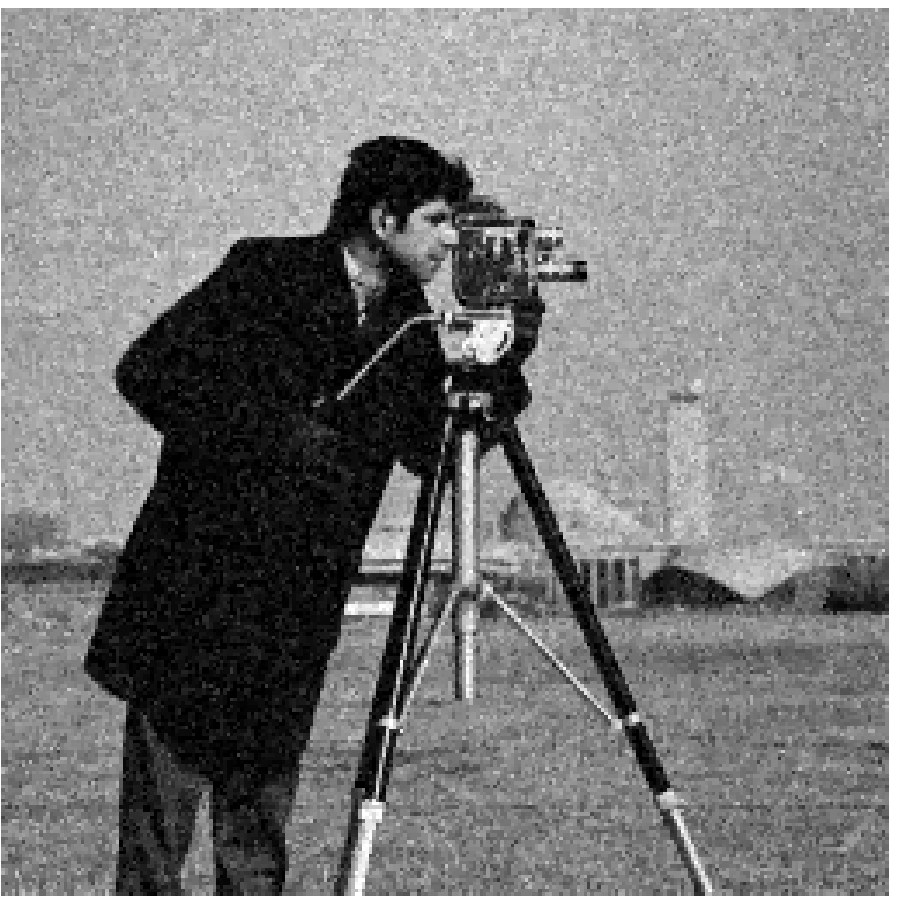}}\hfill
  \subfigure[$\lambda = 10^{-2} \lambda_{\mathrm{bnd}}$]{\includegraphics[height=0.2\linewidth]{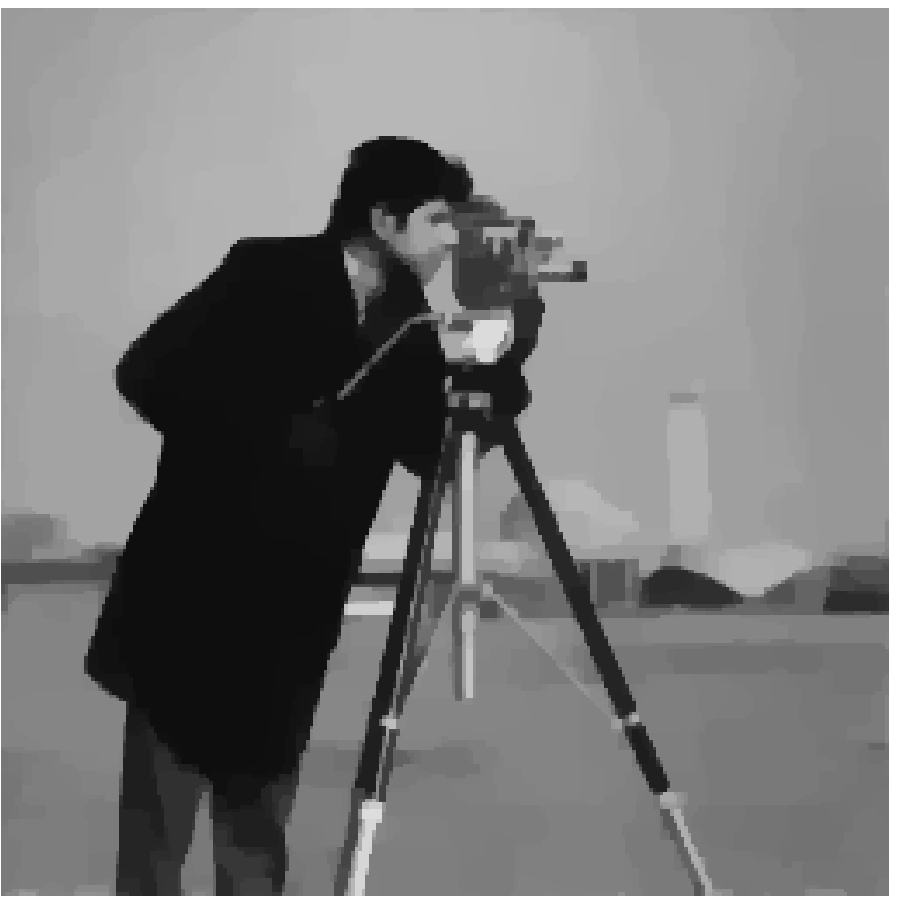}}\hfill
  \subfigure[$\lambda = \lambda_{\mathrm{bnd}}$]{\includegraphics[height=0.2\linewidth]{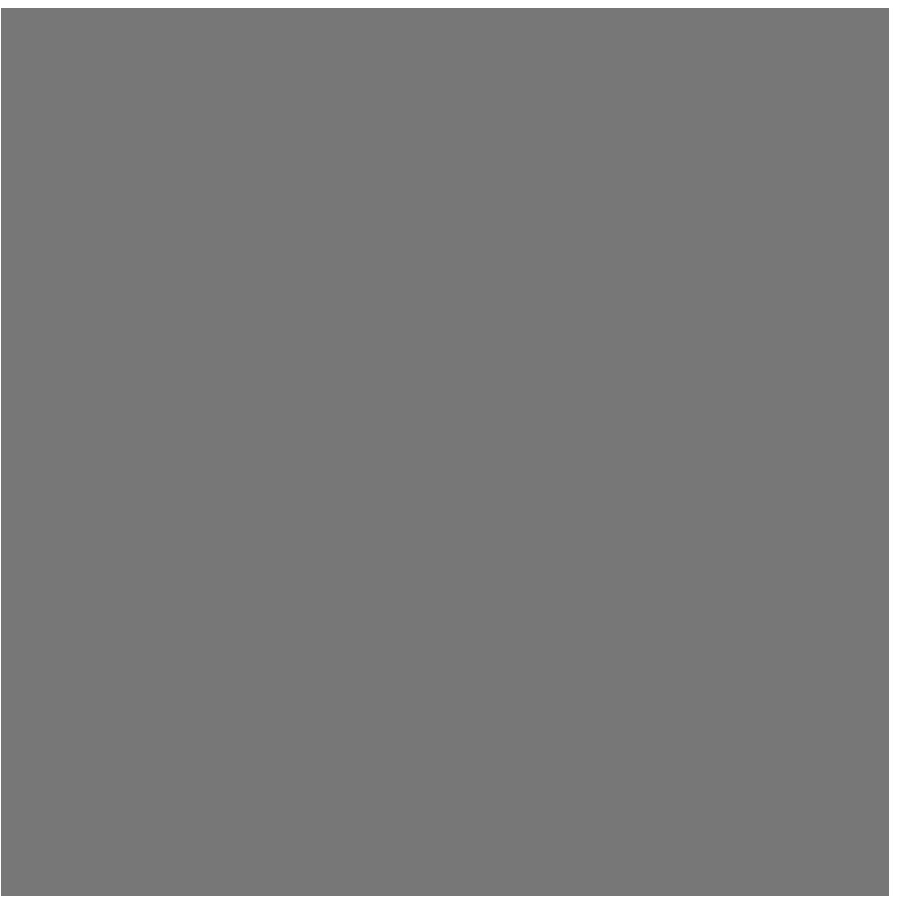}}
  \caption{(a) Evolution of $\norm{\nabla x^\star\,}_\infty$ as a function of $\lambda$.
    (b), (c), (d) Results $x^\star$ of the periodical anisotropic total-variation
    for three different values of $\lambda$.
  }
  \label{fig:tv_2d_periodic_lambda}
  \vspace{1em}
\end{figure*}

Figure~\ref{fig:tv_1d_periodic} and \ref{fig:tv_2d_periodic}
provide illustrations of the computation of
$\lambda_{\max}$ and $\lambda_{\mathrm{bnd}}$
on a 1d signal and a 2d image respectively.
The convolution kernel is a simple triangle wave in the 1d case
but is more complex in the 2d case.
The operator $\diverg \diverg^+$ is in fact the projector
onto the space of zero-mean signals, i.e., $\Ima[\diverg]$.
Figure \ref{fig:tv_2d_periodic_lambda} illustrates the evolution of $x^\star$
with respect to $\lambda$ (computed with the algorithm of \cite{CP}).
Our upper-bound $\lambda_{\mathrm{bnd}}$ (computed in ${\sim}5$ms) appears to be reasonably tight
($\lambda_{\max}$ computed in ${\sim}25$s with \cite{CP} on Problem \eqref{eq:lambda_max}).

Future work will concern the generalization of these results to other
$\ell_1$ analysis regularization and to ill-posed inverse problems.

\bibliographystyle{IEEEtran}
\bibliography{lambdamax}

\end{document}